\documentclass{article}
\usepackage{spconf, amsmath, graphicx}

\usepackage{enumitem}
\usepackage{booktabs}
\usepackage{graphicx}
\usepackage{flushend}
\usepackage{float}
\usepackage{xcolor} 
\setlist{nosep, leftmargin=14pt}

\title{Employing Graph Representations for Cell-level Characterization \\ of Melanoma MELC Samples}

\name{
	\begin{tabular}{c}
		Luis Carlos Rivera Monroy$^{1}$, Leonhard Rist$^{1}$, Martin Eberhardt$^{2}$, Christian Ostalecki$^{2}$, \\ Andreas Baur$^{2}$, Julio Vera$^{2}$, Katharina Breininger$^{3}$, and Andreas Maier$^{1}$
	\end{tabular}
}
\address{$^{1}$ Pattern Recognition Lab, Friedrich-Alexander-Universität Erlangen-Nürnberg, Erlangen, Germany \\
	$^{2}$ Department of Dermatology, Universitätsklinikum Erlangen,  Erlangen, Germany\\
	$^{3}$ Department Artificial Intelligence in Biomedical Engineering, FAU Erlangen-Nürnberg, Erlangen, Germany}

\begin{document}
	\maketitle
	\begin{abstract}
		Histopathology imaging is crucial for the diagnosis and treatment of skin diseases. For this reason, computer-assisted approaches have gained popularity and shown promising results in tasks such as segmentation and classification of skin disorders. However, collecting essential data and sufficiently high-quality annotations is a challenge. This work describes a pipeline that uses suspected melanoma samples that have been characterized using Multi-Epitope-Ligand Cartography (MELC). This cellular level characterization of the tissue is then represented as a graph and finally used to train a graph neural network. This imaging technology, combined with the methodology proposed in this work, achieves a classification accuracy of $87\,\%$, outperforming existing approaches by $10\,\%$.
	\end{abstract}
	
	
	\section{Introduction}
	\label{sec:intro}
	
	A central goal of post-genomic research is to understand the association between a cell or tissue with its specific molecular network. These networks describe specific cellular functions; proteins must be at the proper location, temporal synchronization, and concentration to cooperatively interact with the rest of the network to exert an action \cite{Schubert_2003}. Any molecular network representing cellular functionality obeys hierarchical protein assembly rules. Hence, a protein network-specific architecture can be derived from any tissue \cite{schubert_2010}.
	
	Consequently, proteins with known fluorescent affinity have been used to examine collocations and interactions at a molecular level. Direct fluorescence can only measure up to 17 different proteins \cite{schubert_2008}. Consequently, it is technically unattainable to map sufficient features that depict an accurate protein network. More recently, scanning mass spectrometry has shown promising results boarding this issue, nonetheless, with an inferior spatial resolution \cite{Eckhardt_2013}.
	
	As a result, an imaging methodology that addresses the previously mentioned limitation has gained traction in recent years. The Multi Epitope Ligand Cartography (MELC) is a multiplex imaging technique wherein the chemical responses of tissue are recorded simultaneously as pixels, which enables the mapping of protein networks, the localization of different cell types, and the research of proteome changes derived from diseases \cite{Schubert_2006, Bonnekoh2007, Ruetze_2010}. 
	
	In contrast to other imaging methodologies, MELC can directly represent the hierarchical organization of the cells in a tissue (topology). Moreover, this is medically relevant since the topology of a tissue undergoes constant changes due to migration, differentiation, death and activation status. Characterizing this behaviour provides insight into the development of skin-related diseases, treatment effectiveness, relapse probability, and target therapy design~\cite{krishna_2021}.
	
	Melanoma is the third most common type of skin cancer. However, it is associated with the highest number of deaths from this type of cancer. Traditionally, the prognosis of the disease is carried out by a medical specialist (radiologist or oncologist). The diagnosis will be based on the patient's clinical history and stained histopathology images from the suspected lesion \cite{Chopra_2020}. The specialist will focus on tissue asymmetries, reactivity to some specific agents, lesion growth over time, and the practical knowledge from the speciality taking the task \cite{Chopra_2020}. To help specialists, mathematical models have gained popularity. These computational melanoma models use all clinical data available and images from the sample to find a general description of the lesion and possible forecasts. However, it is still limited by the vast number of melanoma subtypes, the standardization of protocols, and the balance on the detail level of the lesion characterization \cite{Albrecht_2020}.
	In order to find a more generalist approach, graphs have gained traction by following the idea of encoding features from multiple domains and contextualized biological parameters used in traditional mathematical modelling. Knowledge graphs can efficiently and effectively model biological problems, such as protein migration, drug pathway reconstruction, gene ontology, and others \cite{Mohamed_2020}.
	
	\begin{figure*}[htp]	
		\begin{minipage}[]{1.0\linewidth}
			\centering
			\centerline{\includegraphics[scale=1]{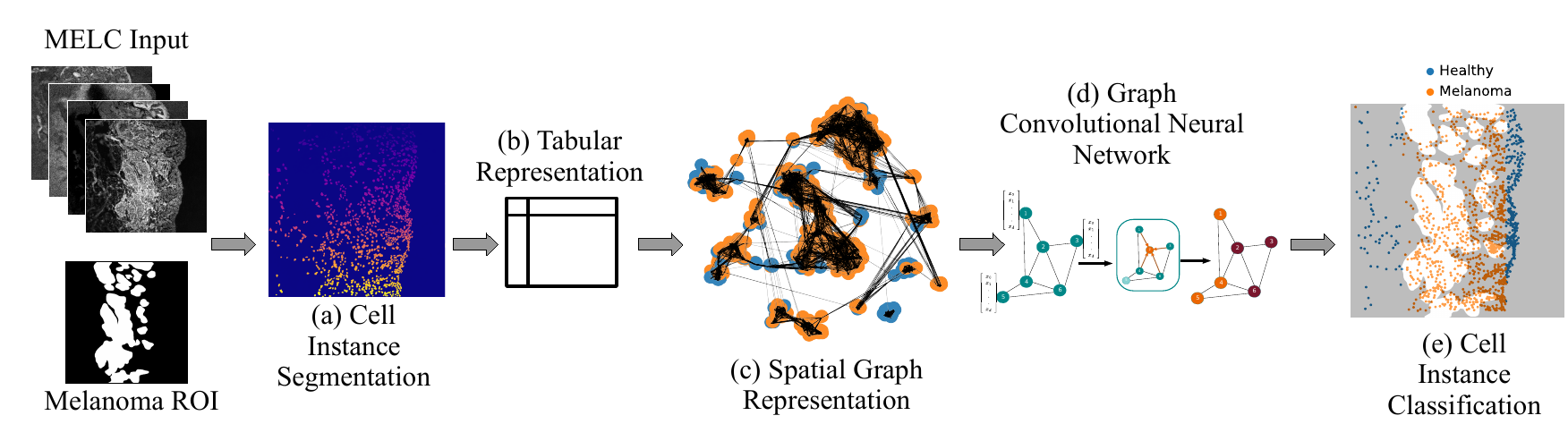}}
			\caption{\textbf{Overview of the proposed methodology.} In the first place, the segmentation of cell instances (a), and second the representation of the stain profiles as tabular data (b). Thirdly, the graph representation of the cells (c) and, finally, the node classification was trained using a graph convolutional neural network (d). The outcome is a binary classification for cell instances on the tissue sample (e).}
		\end{minipage}
		\label{fig:pipeline}
	\end{figure*}
	
	We propose a supervised procedure to classify cells as part of melanoma or a healthy region of skin tissue profiled using MELC. In this work:
	\begin{itemize}
		\item  We propose a pipeline to represent MELC profiles using cell instance segmentation to build cell-wise feature vectors. The detected cells are represented as graph nodes, which allows us to derive a holistic representation of the MELC images.
		\item We investigate two different strategies to connect the graph nodes: feature similarity and spatial proximity. 
		\item We evaluate the performance effect of dimension reduction over the features encoded in the graph.
	\end{itemize}	
	
	\hfill
	\par These approaches are compared to baseline methods such as XGBoost and random forest and evaluated for melanoma cell characterization for 27 patients.
		
	\section{Materials and Methods}
	\label{sec:materials}	

	\subsection{Dataset}	
\begin{figure}[htb]	
		\begin{minipage}[]{1.0\linewidth}
			\centering
			\centerline{\includegraphics[scale=1]{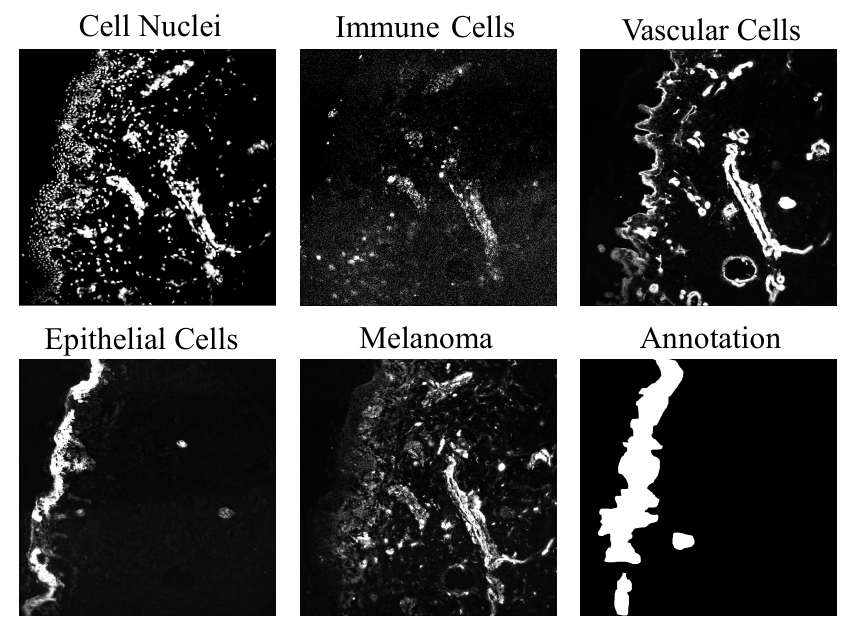}}
		\end{minipage}
		\caption{\textbf{Melanoma MELC sample.} Tissue stained to the five cell sub-types targets and example of staining agent specific for each type. Cell nuclei (Propidium iodide), Immune cells (CD43), Vascular cells (Collagen IV), Epithelial cells (Phospho Connexin), Melanoma associated stain agents (CD63),  and the manual annotation of the melanoma tumour provided by the specialist.}
		\label{fig:dataset}
	\end{figure}

	We evaluate our method on a dataset of suspected melanoma tissue samples for cell-wise classification. The dataset assembled for this work comprised samples from 27 suspected cases, where 20 include positive melanoma cases, and the remaining were healthy tissue. Each of the samples was stained following the MELC protocol. MELC is a microscopy-based imaging technique in which a tissue sample undergoes a repeated cycle of (1) incubation with affinity reagents, (2) staining with fluoroscope-coupled antibodies, (3) immune fluorescence microscopy and (4) soft bleaching of the staining agent. Each sample can be exposed to up to one hundred staining agents following this procedure \cite{Schubert_2006}; each staining produces an image like the examples shown in Figure \ref{fig:dataset}. Each sample was stained using between 80-85 antigens in this dataset. Then, each stain is digitalized as an image of $2018 \times 2018$ pixels (resolution of $0.45\, \mu m /$ pixel). Finally, a medical specialist reviewed each sample, and if a tumour was present, manual segmentation was provided.
	
	This technology is of high relevance due to its capabilities to map protein networks, characterize different sub-groups of cells and track changes in the proteome \cite{Bonnekoh_2008, Berndt_2008}, which presents an opportunity to use this cell-level information to model these representations computationally.
	
	\subsection{Graph Representation}
	Graphs have proven to be capable of depicting complex cellular changes that characterize biological systems. Skin diseases like melanoma are known for their fast and changing behaviour, making earlier diagnosis and proper treatment difficult. Following this premise, we wanted to combine the accurate cellular-level information obtained using the MELC methodology and propose possible ways to represent this data as graphs. First, we use Cellpose \cite{Stringer_2021}, a generalist algorithm, to produce instance segmentation of cells, as shown in Figure  \ref{fig:pipeline}(a). Each of these instances is then profiled for each staining agent to which the sample was exposed. This profile is the average reactivity (brightness) of a cell to each staining agent, which is then depicted as tabular data, as presented in Figure \ref{fig:pipeline}(b).
	
	Second, the next step is the graph representation creation, Figure \ref{fig:pipeline}(c). In this work, we explore two forms of encoding the tissue information into a graph, one in which the nodes (cells) are connected to other nodes via edges based on feature similarity (edges) and spatial proximity. For the first approach, we used the implementation of Wolf et al. \cite{Wolf_2018} in which the size of local neighbourhoods needed to be manually selected (we determined this size to be ten cells, which empirically showed stable performance along the different methods); Kendall's $\tau$ is used as the similarity criteria. We followed the implementation proposed by Palla et al. \cite{palla_squidpy_2022} for the second representation. This representation uses the spatial coordinates of the cells as the primary input. To maintain consistency between the approaches, we also defined the number of connected nodes as ten.
	
	\subsection{Dimensionality Reduction}
	Neighbourhood graph-based approaches have shown to be vital to initialization sensitivity, and suboptimal local minima optimization is usually present on typical graphs. However, this comes with a substantial computational cost; by definition, a neighbourhood graph construction is quadratic in the number of data points \cite{Gorban_2008}. To approach this issue, we analyze the effect of two-dimensionality reduction algorithms that are relevant when used in the single-cell analysis of multi-array histopathology images, outperforming more traditional methods like PCA \cite{Wu_2019, Wang_2021, Do_2021}. First, Uniform Manifold Approximation and Projection UMAP \cite{McInnes_2018} is a manifold learning technique based on Riemannian and algebraic geometry with no computational restrictions on the embedding dimension. Secondly, tSNE \cite{Maaten_2008} is used as a non-parametric conditionality reduction technique representing crowded points and multi-scale structures. Both of these schemes will redefine the feature representation at each node.
	
	\subsection{Graph Neural Network (GNN)}	
	The next step in our pipeline is the node (cell) classification using a GNN as presented in Figure \ref{fig:pipeline}(d). For this task, we used the Graph Random Neural Network (GRAND) \cite{Feng_2020} architecture. This architecture presents random propagation, a strategy in which multiple graph augmentations are generated stochastically. This type of augmentation is consistent because neighbour nodes tend to have similar labels. Therefore, the information dropped in one augmentation can be compensated by the neighbourhood. Consequently, the model achieves better generalization and, compared with other state-of-the-art networks, better robustness and resistance to the over-smoothing effect \cite{Feng_2020}.
	
	\subsection{Experimental Setup and Metrics}
	For the performance evaluation, the dataset was divided following a $70\,\% / 10\,\% / 20\,\%$ split for training, validation, and testing, respectively. Every split had at least one sample of healthy and melanoma tissue. The hyper-parameter optimization was performed on the train split following a Bayesian search scheme and limited by 50 iterations without improvement, which empirically showed to be a reasonable restriction for all the methods.
	
	\textbf{Baseline models:} As a point of reference to compare the classification performance of the graphs we used XGBoost and Random Forest. We chose these two methods because both are standard baselines for machine learning tasks using tabular data as input. We used the implementation available on Scikit-learn \cite{scikit-learn}. 
	
	\textbf{GRAND:} For the training of the GNN, we used the implementation that the authors made available. We trained using a GPU (Quadro RTX 4000). The parameter optimization showed that the following setup produces the best performance on the train split: A drop-node rate probability of $0.45$, an initial learning rate of $0.003$, an early stop tolerance $100$ rounds, the number of epochs to $2500$, and remaining parameters set to default.
	
	The metrics used to evaluate the node/cell classification performance were the standard accuracy, F1-Score and the Area under the Receiver Operator Characteristic curve (AUROC).
	
	\section{Experimental Results}
	\label{sec:mresults}	
	
	Table \ref{table:results} summarizes the prediction accuracies of node classification. We evaluated the performance of different embeddings and the impact that some dimensionality reduction schemes will have on the performance of melanoma cell classification. 
	
	\begin{table}[h]
		\caption{\textbf{Performance on melanoma dataset.} Results report of the different data embedding type and data reduction schemes for the node classification. Best scores are marked in bold.}
		\resizebox{\columnwidth}{!}{%
			\begin{tabular}{@{}c|cccc@{}}
				\toprule
				\textbf{Embedding} & \textbf{Dimension Reduction} & \textbf{Accuracy} & \textbf{F1-Score} & \textbf{AUROC} \\ \midrule
				\textbf{Tabular - XGBoost} & - & 0.71 & 0.75 & 0.78 \\ \midrule
				\textbf{Tabular - Random Forest} & - & 0.74 & 0.70 & 0.66 \\ \midrule
				\textbf{} & - & 0.65 & 0.75 & 0.73 \\
				\textbf{Graph - Feature Neighbourhood} & UMAP & 0.83 & 0.82 & \textbf{0.89} \\
				\textbf{} & tSNE & 0.60 & 0.77 & 0.71 \\ \midrule
				\textbf{} & - & 0.79 & 0.69 & 0.70 \\
				\textbf{Graph - Spatial Neighbourhood} & UMAP & \textbf{0.87} & \textbf{0.90} & 0.88 \\
				\multicolumn{1}{l|}{} & tSNE & 0.78 & 0.70 & 0.71 \\ \bottomrule
			\end{tabular}%
			\label{table:results}
		}
	\end{table}
	
	\section{Discussion}
	
	From Table \ref{table:results}, we can observe that the graph representation that considers the spatial location of cells achieves consistently better performance than the one that only encodes the reactivity profiles. It is also possible to observe similar performance between the methods that rely only on the tabular profiles and the feature graph. This is expected because these three encode the same type of information.

	When it comes to the performance of dimensionality reduction on graphs, UMAP produces the best performance. As described by the authors, this representation favours the description and quantification of spatial patterns, and cellular neighbourhoods across a sample \cite{palla_squidpy_2022}, which is coherent with the expected behaviour of biological systems. While tSNE strongly favours the preservation of the feature correlation. Even though there is a performance improvement when using these methods, it is expected that the most significant impact is on the computational resources side, which will be reflected in training time and memory usage, for example. 
	
	We observe that the best overall performance is achieved when combining the spatial neighbourhood graph with UMAP, which produces an improvement of around $10\,\%$ compared with the other settings. This result reflects the importance of modelling spatial connectivity on biological systems and the advantage of including dimensionality reduction as a preprocessing step.
	
	\section{Conclusions and Future Work}
	\label{sec:conclusions}
	
	This paper proposes a pipeline for cell instance classification on melanoma histopathology samples. We study the viability of combining the representation of multi-array images (MELC stains) into graphs and using graph neural networks to train a classifier. First, we explored two ways of representing these profiles in graphs, the differentiable factor being the connectivity criteria, one by similarity and the other by spatial proximity. Furthermore, we evaluated the performance of this representation using one of the state of the art architectures for node classification (GRAND). Even though state-of-the-art approaches, such as XGBoost, perform vastly well on tabular data, these methods need to incorporate spatial biological information and therefore are limited when representing biological systems. Additionally, this work explored the effect of some dimensionality reduction schemes commonly used in cell-level analysis.
	Our experiments indicated that the graph representation in which the spatial connectivity and the UMAP dimensionality reduction are combined produces the best result. 
	Further research will focus on studying the impact of adaptative stain-specific normalisation techniques and the size of neighbourhoods in the overall classification capabilities of pipelines like the one discussed in this work. 
	
	\section{Compliance with ethical standards}
	\label{sec:ethics}
	
	The human tissue examinations are performed in close consultation and cooperation with the clinic's histopathologists. The samples obtained for MELC technology are archived in an officially established (by the Institute of Medical Documentation in Erlangen, Germany) tissue bank. A patient information sheet or informed consent form was developed in agreement with the ethics committee to collect the samples. After consultation with the ethics committee, MELC analysis is equivalent to extended immunofluorescence diagnostics as routinely used for histological diagnostics. A corresponding ethics application has been approved for this project (Application no.: 43\_18 B).	
	
	\section{Acknowledgments}
	\label{sec:acknowledgments}
	
	Katharina Breininger gratefully acknowledges support by Dhip campus - Bavarian aim in form of a faculty endowment.
	
	\bibliographystyle{IEEEbib}
	\bibliography{refs.bib}

\end{document}